\documentclass[sigconf]{aamas}  

\usepackage{booktabs}
\usepackage{graphicx}
\usepackage{times}
\usepackage{latexsym}
\usepackage{epsfig}
\usepackage{algorithm, algpseudocode}
\usepackage{multirow}
\usepackage{makecell}
\usepackage{color}
\usepackage{subfigure}
\usepackage{url}
\usepackage{mathrsfs}
\usepackage{amsmath}
\usepackage{url}

\newcommand{\secref}[1]{Section \ref{#1}}

\newcommand{\bi}[1]{\textbf{\textit{#1}}}

\setcopyright{ifaamas}  
\acmDOI{doi}  
\acmISBN{}  
\acmConference[AAMAS'19]{Proc.\@ of the 18th International Conference on Autonomous Agents and Multiagent Systems (AAMAS 2019), N.~Agmon, M.~E.~Taylor, E.~Elkind, M.~Veloso (eds.)}{May 2019}{Montreal, Canada}  
\acmYear{2019}  
\copyrightyear{2019}  
\acmPrice{}  

\begin{document}

\title{A Multi-task Selected Learning Approach for Solving \\ 3D Flexible Bin Packing Problem}  

\subtitle{Industrial Applications Track}


\author{Lu Duan}  
\affiliation{
  \institution{Artificial Intelligence Department, Zhejiang Cainiao Supply Chain Management Co., Ltd}
}
\email{ duanlu.dl@cainiao.com}

\author{Haoyuan Hu}  
\affiliation{
	\institution{Artificial Intelligence Department, Zhejiang Cainiao Supply Chain Management Co., Ltd}
}
\email{ haoyuan.huhy@cainiao.com}

\author{Yu Qian}  
\affiliation{
	\institution{Artificial Intelligence Department, Zhejiang Cainiao Supply Chain Management Co., Ltd}
}
\email{ qianyu.qy@alibaba-inc.com}

\author{Yu Gong}  
\affiliation{
	\institution{Search Algorithm Team, Alibaba Group}
}
\email{ gongyu.gy@alibaba-inc.com}

\author{Xiaodong Zhang}  
\affiliation{
	\institution{Artificial Intelligence Department, Zhejiang Cainiao Supply Chain Management Co., Ltd}
}
\email{yuhui.zxd@cainiao.com}

\author{Jiangwen Wei}  
\affiliation{
	\institution{Artificial Intelligence Department, Zhejiang Cainiao Supply Chain Management Co., Ltd}
}
\email{jiangwen.wjw@alibaba-inc.com}

\author{Yinghui Xu}  
\affiliation{
	\institution{Artificial Intelligence Department, Zhejiang Cainiao Supply Chain Management Co., Ltd}
}
\email{renji.xyh@taobao.com}

\begin{abstract}  
A 3D flexible bin packing problem (3D-FBPP) arises from the process of warehouse packing in e-commerce. An online customer's order usually contains several items and needs to be packed as a whole before shipping. In particular, $5\%$ of tens of millions of packages are using plastic wrapping as outer packaging every day, which brings pressure on the plastic surface minimization to save traditional logistics costs. Because of the huge practical significance, we focus on the issue of packing cuboid-shaped items orthogonally into a least-surface-area bin. The existing heuristic methods for classic 3D bin packing don't work well for this particular NP-hard problem and designing a good problem-specific heuristic is non-trivial. 
In this paper, rather than designing heuristics, we propose a novel multi-task 
framework based on Selected Learning to learn a heuristic-like policy that generates the sequence and orientations of items to be packed simultaneously. Through comprehensive experiments on a large scale real-world transaction order dataset and online AB tests, we show: 1) our selected learning method trades off the imbalance and correlation among the tasks and significantly outperforms the single task Pointer Network and the multi-task network without selected learning; 2) our method obtains an average $5.47\%$ cost reduction than the well-designed greedy algorithm which is previously used in our online production system.
\end{abstract}

%


\maketitle

\section{Introduction}
\label{sec:intro}
With the vigorous development of e-commerce, related logistic costs have risen to about 15\% of China's GDP\footnote{\url{https://www.alizila.com/jack-ma-alibaba-bets-big-on-logistics/}}, attracting more and more attention. As the largest e-commerce platform in China, Taobao (taobao.com) serves over six hundred million active users and reducing costs is often the top priority of its logistics system. There are many methods that can help cut down logistic costs, e.g, reducing the packing costs. 
Tens of millions of packages are sent to customers every day, $5\%$ of the which are prepared using plastic wrappers. In order to reduce the cost of plastic packing materials, warehouse operation prefers to pack the items as a whole in a way that minimize the wrapping surface area. In this paper, we formalize this real-world scenario into a specific variant of the classical three-dimensional bin packing problem (3D-BPP) named 3D flexible bin packing problem (3D-FBPP). The 3D-FBPP is to seek the best way of packing a given set of cuboid-shaped items into a flexible rectangular bin in such a way that the surface area of the bin is minimized.  


The 3D-FBPP, which is the focus of this work, is a new problem and has been barely studied. However, the 3D-BPP, a similar research direction of 3D-FBPP, has been extensively studied in the field of operational research (OR) during last decades \cite{wascher2007improved}. 
As a strongly NP-hard problem \cite{martello2000three}, the traditional approaches to tackling 3D-BPP have two main flavors: exact algorithms \cite{chen1995analytical,martello2007algorithm} and heuristics \cite{gonccalves2013biased,lodi2002heuristic}. While exact algorithm provides optimal solution, it usually needs huge amount of time to solve modern size instances. Heuristic algorithm, which cannot guarantee optimal, is capable to give acceptable solutions with much less computational effort. 

Typically, to achieve good results and guarantee the computational performance, traditional exact algorithms and heuristics require large amount of expertise or experience to design specific search strategies for different types of problems. In recent years, 
there has been some seminal work on using deep architectures to learn heuristics for combinatorial problems, e.g, Travel Salesman Problem (TSP) \cite{bello2016neural,vinyals2015pointer,kool2018attention}, Vehicle Routing Problem (VRP) \cite{nazari2018deep}. These advances justify the renewed interest in the application of machine learning (ML) to optimization problems. Motivated by these advances, this work combines reinforcement learning (RL) and supervised learning (SL) to parameterize the policy to obtain a stronger heuristic algorithm. 

In 3D-FBPP problem, the smaller objective (i.e. the surface area that can pack all the items of a order) is better, implying a better packing solution. To minimize the objective, a natural way is to decompose the problem into three interdependent decisions \cite{gonccalves2013biased,li2014genetic}: 1) decide the sequence to place these items; 2) decide the place orientation of the selected item; 3) decide the spatial location. These three decisions can be considered as learning tasks, however, in our proposed method, we concentrate on the first two tasks, using RL and SL, respectively. Specially, we adopt an intra-attention mechanism to address the repeating item problem for the first sequence decision task. To learn a more competitive orientation strategy, we adopt the idea of hill-climbing algorithm, which will always keep the current best sampled solution as the ground-truth and make incremental change on it. Regardless of spatial location, the sequence of packing the items into the bin will influence the orientations of each item and vice versa, so the two tasks are correlated. Meanwhile, with $n$ items, the choice of orientations is $6^n$ and the choice of sequence is $n!$, so the two tasks are difficulty-unbalanced in learning process. Inspired by multi-task learning, 
we adopt a new type of training mode named Multi-task Selected Learning (MTSL) to utilize correlation and mitigate imbalance mentioned above. MTSL is under the following settings: each subtask and both of them are treated as a training task. In MTSL, we select one kind of the training tasks according to a probability distribution, which will decay after several training steps ( \secref{sec:mtsl-train}). 

In this paper, we present a heuristic-like policy learned by a neural model and 
quantitative experiments are designed and conducted to demonstrate effectiveness of this policy. The contributions of this paper are summarized below.
\begin{itemize}	
	\item This is a first and successful attempt to define and solve the real-world problem of 3D flexible Bin Packing ( \secref{sec:problem}). We collect and will open source a large-scale real-world transaction order dataset (LRTOD) ( \secref{sec:data}). By modeling packing operation as a sequential decision-making problem, the proposed method naturally falls into the category of reinforcement learning and it is one of the first applications of reinforcement learning in large-scale real-time systems.
	\item We use an intra-attention mechanism to tackle the sequential decision problems during the optimization, which considers items that have already been generated by the decoder.  
	\item We propose a multi-task framework based on Selected Learning, which can significantly utilize correlation and mitigate imbalance among training tasks. Based on this framework, packing sequence and orientations can be conducted at the same time. 
	\item We achieves $6.16\%$, $9.66\%$, $8.25\%$ improvement than the greedy heuristic algorithm designed for the 3D-FBPP in BIN8, BIN10 and BIN12 and an average $5.47\%$ cost reduction on online AB tests. Numerical results also demonstrate that our selected learning method significantly outperforms the single task Pointer Network and the multi-task network without selected learning.
\end{itemize}


\section{Related Work}
\label{sec:related}
\noindent
\subsection{Neural encoder-decoder sequence models}
Neural encoder-decoder models have provided remarkable results in Neural language Process (NLP) applications such as machine translation \cite{sutskever2014sequence}, question answering \cite{hermann2015teaching}, image caption \cite{xu2015show} and summarization \cite{chopra2016abstractive}. These models encode an input sequence into a fixed vector by a recurrent neural networks (RNN), and decode a new output sequence from that vector using another RNN. Attention mechanism, which is used to augment neural networks, contributes a lot in areas such as machine translation (\cite{bahdanau2014neural}) and abstractive summarization \cite{paulus2017deep}. In \cite{paulus2017deep}, intra-attention mechanism, which considers words that have already been generated by the decoder, is proposed to address the repeating phrase problem in abstractive summarization. The repeating problem also exists in our 3D-FBPP, since an item cannot be packed into a bin more than twice. In this study, we adopt the special intra-attention mechanism for producing ``better'' output.

\noindent
\subsection{Machine learning for combinatorial optimization} 
The application of ML to discrete optimization problems can date back to the 1980's and 1990's \cite{smith1999neural}. However, very limited success is ultimately achieved and the area of research is left nearly inactive at the beginning of this century. As a result, these NP-hard problems have traditionally been solved using heuristic methods \cite{silver2004overview,bortfeldt2007heuristic}. Currently, the related work is mainly focused on three areas: learning to search, supervised learning and reinforcement learning. To obtain a strong adaptive heuristics, learning to search algorithm\cite{xia2018learning} adopts a multi-armed bandits framework to apply various variable heuristics automatically. Supervised learning method \cite{vinyals2015pointer} is a first successful attempt to solve a combinatorial optimization problem by using recent advances in artificial intelligence. In this work, a specific attention mechanism named Pointer Net motivated by the neural encoder-decoder sequence model is proposed to tackle TSP. Reinforcement learning aims to transform the discrete optimization problems into sequential decision problems. 
Base on Pointer Network, \cite{bello2016neural} develops a neural combinatorial optimization framework with RL, which solves some classical problems, such as TSP and Knapsack Problem. Similar works using architecture like Pointer Network can also be seen in \cite{nazari2018deep,kool2018attention}. On a related topic, \cite{khalil2017learning} solves optimization problems over graphs using a graph embedding structure and a greedy algorithm learned by deep Q-learning (DQN). However, there are still some work about one-dimensional bin packing problem. \cite{mao2017small,sim2015lifelong} try to select or generate the best heuristic that can
generate a better quality bin packing solution. Our work focuses on the 3D-FBPP and the main difference between our work and the previous work (e.g, TSP, VRP) is that our work consists of several tasks that are imbalanced and correlated, which brings a grave challenge.

\section{Problem Statement}
\label{sec:problem}
In real online e-commerce delivery phase, a large number of packages are wrapped with plastic materials as outer packaging before shipping. The goal of this scenario is to minimize the packing materials per order by generating a better packing solution. 
After stacking all the items, the plastic material can be made into a rectangular-shaped bin-like outer wrapping in warehouse . As the result, the cost of this material is directly proportional to the surface area of the rectangular-shaped bin. In this case, minimizing the surface area for the bin would bring huge economic benefits for traditional logistics.

The mathematical formulation of 3D-FBPP is shown below. Given a set of cuboid-shaped items and each item $i$ is characterized by length ($l_i$), width ($w_i$) and height ($h_i$). Our target is to find the least-surface-area bin that can pack all items. 
Generally, we use $(x_i, y_i, z_i)$ to denote the front-left-bottom (FLB) coordinate of item $i$ and assume that FLB corner of the bin is $(0,0,0)$.  
To ensure that there is no overlap, binary variables $s_{ij}$, $u_{ij}$, $b_{ij}$ are defined to indicate the placement of items $i$ to each item $j$. $s_{ij}$, $u_{ij}$, $b_{ij}$ is equal to 1 if items $i$  is  left of, under of, back of item $j$ respectively; otherwise 0. The variable ${\delta}_{i1}$(resp. ${\delta}_{i2}$, ${\delta}_{i3}$, ${\delta}_{i4}$, ${\delta}_{i5}$, ${\delta}_{i6}$) is equal to 1 if the orientation of item $i$ is front -up (resp. front-down, side-up, side-down, bottom-up, bottom-down). Our aim is to find a least-surface-area bin with size $(L,W,H)$, where $L$, $W$ and $H$ is the length, width and height of the bin respectively. 

Based on the descriptions of problem and notations, the mathematical formulation for the 3D-FBPP is followed by \cite{hifi2010linear}:
\begin{eqnarray*}
\min\,\,  L \cdot W + L \cdot H + W \cdot H\\ 
\end{eqnarray*}
subject to the following set of constraints:
\begin{eqnarray}
 &s_{ij}+ u_{ij} + b_{ij}  =1  \\ 
 &{\delta}_{i1}+{\delta}_{i2}+{\delta}_{i3}+{\delta}_{i4}+ {\delta}_{i5} + {\delta}_{i6} =1  \\ 
 &x_i - x_j + L\cdot s_{ij} \le L-\hat{l_i}                             \\ 
 &y_i - y_j + W \cdot b_{ij} \le W - \hat{w_i}                          \\ 
 &z_i - z_j + H \cdot u_{ij} \le  H - \hat{h_i}     \\                                                                               
 &0 \le x_i \le L- \hat{l_i}                                     \\ 
 &0 \le y_i \le W - \hat{w_i}                                 \\ 
 &0 \le z_i \le H - \hat{h_i}                                   \\ 
 &\hat{l_i} = {\delta}_{i1}   l_i + {\delta}_{i2}  l_i + {\delta}_{i3}  w_i + {\delta}_{i4}  w_i + {\delta}_{i5}  h_i + {\delta}_{i6}  h_i \\  
 &\hat{w_i} = {\delta}_{i1}  w_i +{\delta}_{i2}  h_i +{\delta}_{i3}  l_i +{\delta}_{i4}  h_i +{\delta}_{i5}  l_i +{\delta}_{i6}  w_i \\ 
 &\hat{h_i} =  {\delta}_{i1}  h_i +{\delta}_{i2}  w_i +{\delta}_{i3}  h_i +{\delta}_{i4}  l_i + {\delta}_{i5}  w_i +{\delta}_{i6}  l_i \\
 &s_{ij} , u_{ij},  b_{ij}  \in \{0,1\}       \\
 &{\delta}_{i1}, {\delta}_{i2}, {\delta}_{i3}, {\delta}_{i4}, {\delta}_{i5}, {\delta}_{i6} \in \{0,1\}                                                    
\end{eqnarray}	

\begin{figure}[!htb]
	\centering
	\epsfig{file=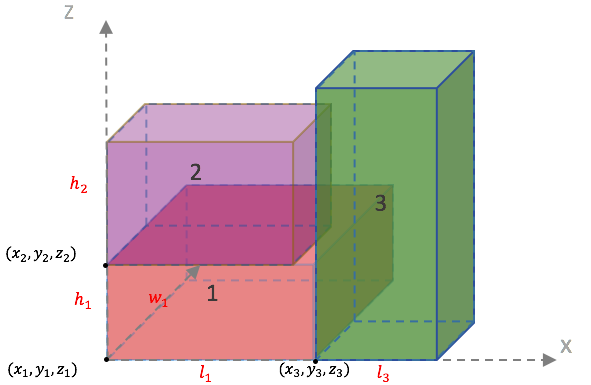, width=0.6\columnwidth}
	\caption{Illustration of non overlapping constraint: item 1 is under item 2 and this means $u_{1,2}=1$ and $z_1 + h_1 <= z_2$, which is constraint (5); item 1 is in the left of item 3 and this means $s_{1,3}=1$ and $x_1 + l_1 <= x_3$, which is constraint (3). }
	\label{fig:problem-illu}
\end{figure}

Constraints $(9)-(11)$ denote the actual length, width, height of item $i$ after orientating it. Constraints $(1)-(5)$ are used to guarantee there is no overlap between two packed items while constraints $(6)-(8)$ are used to guarantee the item will not be put outside the bin. Figure \ref{fig:problem-illu} explains the non-overlapping constraints in the problem definition.

\section{Multi-task Selected Learning}
\label{sec:model}

In this section, we describe our multi-task selected learning approach and implementation details for solving the 3D-FBPP. 
\begin{figure}[h]
	\centering
	\epsfig{file=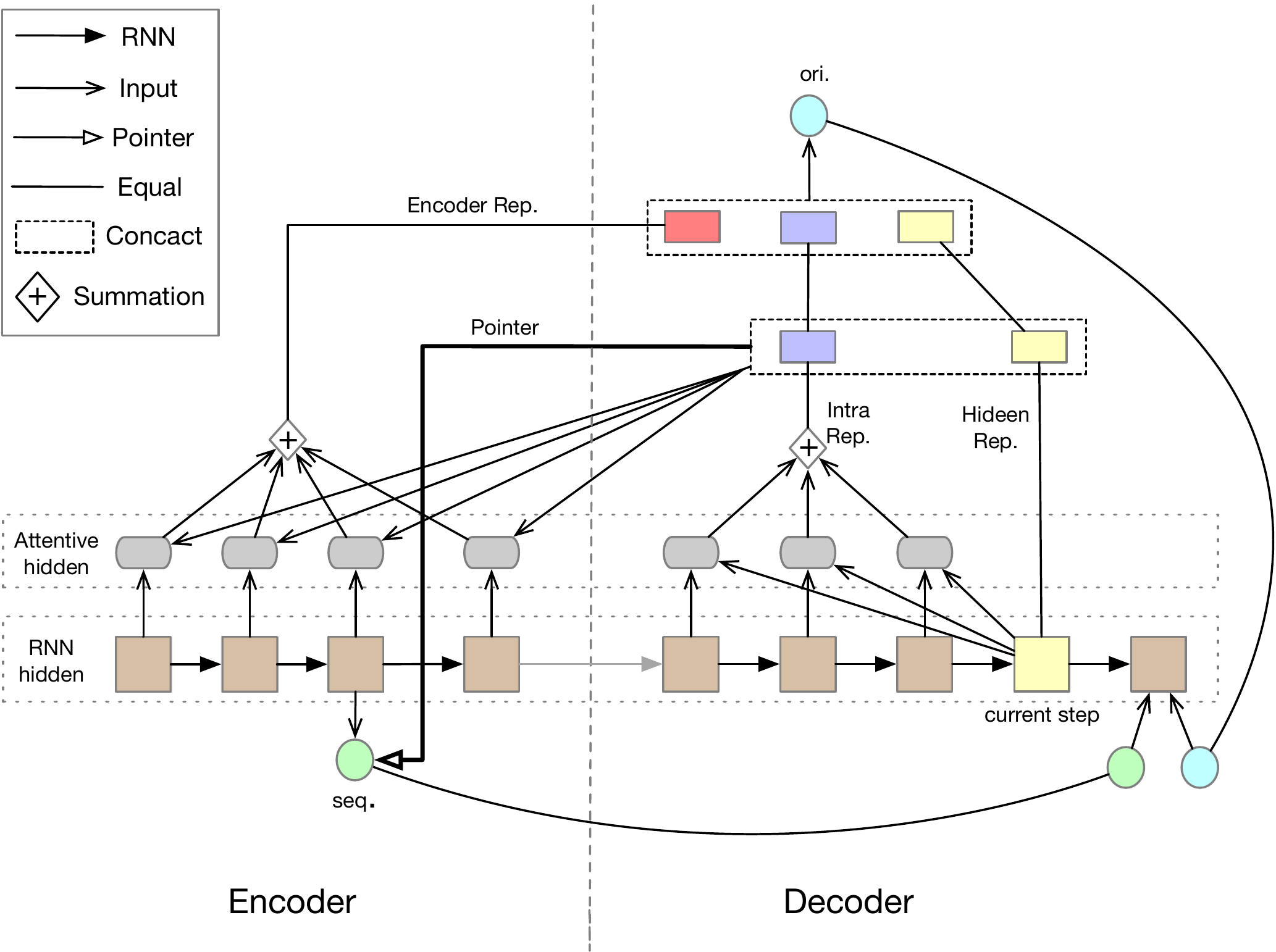, width=1.0\columnwidth}
	\caption{Architecture for multi-task 3D-FBPP networks. Rep.,ori. and seq. is short for representation, orientation output and sequence output, respectively. }
	\label{fig:architecture-model}
\end{figure}

\begin{figure}[h]
	\centering
	\epsfig{file=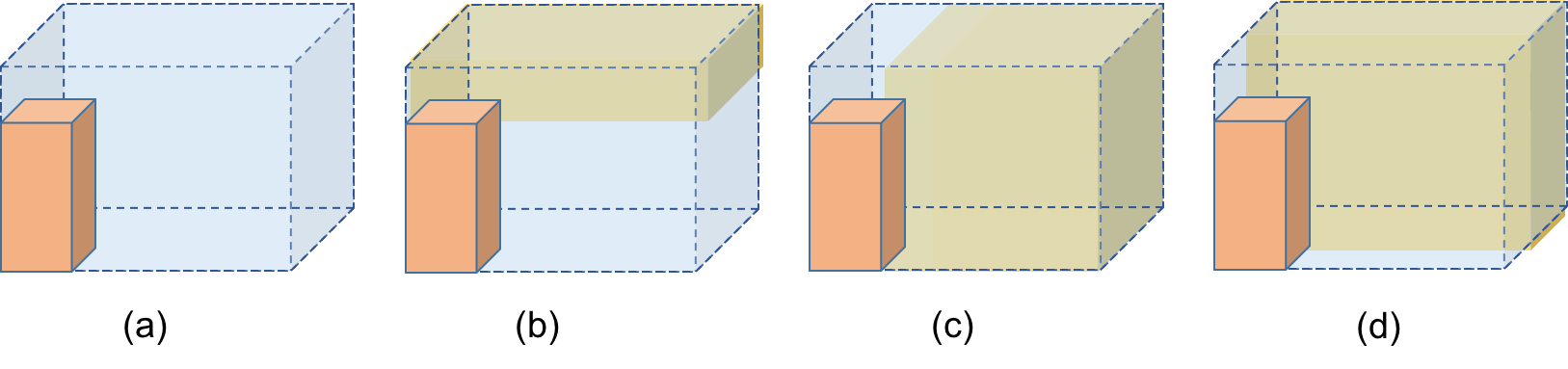, width=0.85\columnwidth}
	\caption{Example of empty-maximal-space. a) item packed in the bin ; b), c), d) three newly generated maximal-spaces shown in yellow. }
	\label{fig:freespace}
\end{figure}

Basically, the procedure of solving the 3D-FBPP can be divided into three related tasks: order of items to be packed (\emph{Sequence Generation}), orientation of each item in the bin (\emph{Orientation Generation}) and front-left-bottom coordinate of each item (\emph{Spatial Location Selection}). Least Waste Space Criteria (LWSC), a heuristic greedy algorithm currently being used in our online production system, inserts the node (item, orientation, spatial location) with least increased surface area to a partial solution. \emph{Orientation Generation} has $3! = 6$ different choices (Suppose an item has 3 sides \emph{a,b,c}; consequently, there are 6 different orientations that each corresponds to one of the following rotated dimensions for this item: \emph{a-b-c}, \emph{a-c-b}, \emph{b-a-c}, \emph{b-c-a}, \emph{c-a-b}, \emph{c-b-a}). The management of the feasible spatial locations is based on a list of empty-maximal-spaces (EMSs) as described in \cite{lai1997developing}. 
Figure \ref{fig:freespace} depicts an example of the generation process of EMSs. In this example we assume that we have one item to be packed in a bin and three new EMSs shown in yellow results from the intersection of the item with the initial empty-maximal-space (empty bin). Each time after an item is placed in an empty-maximal-space, its corresponding new empty maximal-spaces are generated. As a result, during the packing procedure, candidates of EMSs highly depend on the previously placement of items and hard to incorporate with existing machine learning method, such as neural network. 

In this work, we disregard the spatial location selection strategy and greedily choose the empty-maximal-space according to LSWC. Therefore, we concentrate on the sequence and orientation generation tasks. In order to leverage useful information contained in these two related tasks to help improve the generalization performance of all the tasks, we propose a multi-task framework based on selected learning. Sequence task is a sequential decision problem, whereas the orientation task is a classification one. Since these two correlated tasks work together to benefit each other, we prefer to use a parameter-shared framework instead of training them separately.

Recently, pointer network (Ptr) \cite{bahdanau2014neural}, whose output may be expressed as a sequence of indexes referring to elements of the input sequence, has been widely used to learn a heuristic algorithm for sequential decision operation research problem like TSP. In this paper, particular to our multi-task setting, we make some adjustments to meet special demand. The proposed network architecture implicitly separates the representation of sequence decision and orientation prediction. The specific architecture consists of two streams that represent the sequence pointer and orientation predict functions, while sharing a common encoder-decoder learning module. The overall architecture of our method is shown in Figure \ref{fig:architecture-model}. Because different tasks have different learning difficulties and interact with each other, we propose a selected learning mechanism to improve different tasks separately in each round to keep them dynamic balance in training step.

Formally, a 3D-FBPP instance with $n$ items can be denoted as $\bi{x} = \{x_i=(l_i, w_i, h_i)\}_{i=1}^n$, where $l_i$, $w_i$ and $h_i$ represents the length, width and height of item $i$ respectively. The solution of this problem is a sequence of triplets $\{(s_i, o_i, f_i)\}_{i=1}^n$ which must meet the constraints described in Section 3, where $s_i$, $o_i$ and $f_i$ represent the item, orientation and empty-maximal-space to be placed in step $i$ during packing. Notably, $s_i$ and $o_i$ are produced by our model while $f_i$ is calculated by the greedy strategy LWSC mentioned above.  
Our model reads the sequence of input tokens (items \bi{x}) with a Long
Short-Term Memory (LSTM) [16] encoder and decodes
the sequence of output tokens (items \bi{s} and orientation \bi{o}). During decoding steps, the input of decoder cell for time-step $t$ contains two parts denoted as $y_t=(s_{t-1}, o_{t-1})$. Specially, the packing sequence $\bi{s} =\{s_1,s_2,...,s_n\}$ is a permutation of the input items $\bi{x}$. 

\subsection{Sequence Task}
\label{sec:sequence_task}
Due to the particularity of packing operation, it's not allowed to pack the same item repeatedly. The way to prevent the sequence from containing the same item 
twice in the previous work \cite{bahdanau2014neural} introduces a hard constraint and probability distribution of next items to be packed is independent of the already generated item subsequence.
However, taking previously decoded items into consideration means the network can have a priori knowledge to try to avoid repetition to some extend. Moreover, incorporating the information of previous decoding steps into the decoder can help network to generate more reasonable and structured pointers. To achieve this, we introduce an intra-attention mechanism \cite{paulus2017deep}, which is first proposed to solve combinational optimization problem. 

For notation purposes, let us define encoder and decoder hidden states as $h_{i}^{e}$ and $h_{t}^{d}$ respectively. 
Here $h_{i}^{e}$ and $h_{t}^{d}$ are computed from the embedding vector of $x_{i}$ and $y_{t}$ respectively.
We define $attn_{tj}^{d}$ as the intra-attention weight of the previous hidden states $h_{j}^{d}$ at decoding step $t$:
\begin{eqnarray*}\label{1}
	&attn_{tj}^{d} = softmax({v_{1}}^{T}tanh(W_{1}h_{j}^{d} + W_{2}h_{t}^{d})), j \in \{1,\dots,t-1\},
\end{eqnarray*}
where $W_{1}$, $W_{2}$ and $v_{1}$ are trainable parameters for intra-attention.
Thus, the intra-attention feature $h^{intra}_{t}$ of the current decoding item $s_{t}$ is calculated by summing up the previous decoded hidden states based on the computed intra-attention weights, for $t > 1$: 
\begin{eqnarray*}\label{2}
	&h^{intra}_{t} = \sum_{j=1}^{t-1}{attn_{tj}^{d}h_{j}^{d}},	
\end{eqnarray*}

Especially for the first decoder cell, we set $h^{intra}_{1}$ to a vector of zeros since the already generated sequence is empty.

In previous deep reinforcement learning method for combinatorial optimization problems, the pointer mechanism considers the encoder attention weights as the probability distribution to copy the input item. In this work, we get the final attention weights $p(s_{t})$ by integrating intra-attention feature at decoder step. In addition, $u_j^t$ is intermediate result which will be normalized by softmax to be the ``attention'' mask over the inputs.
\begin{eqnarray*}\label{3}
	&u_{j}^{t} = {v_{2}}^{T}tanh(W_{3}h_{j}^{e} + W_{4}h_{t}^{d} + W_{5}h^{intra}_{t}) \nonumber, \\
	&p(s_{t}) = softmax(u^{t}), \quad j\in\{1,2,...,n\},	 
\end{eqnarray*}

where $W_{3}$, $W_{4}$, $W_{5}$ and $v_{2}$ are trainable parameters for pointer networks.
We refer all parameters for sequence task as $\theta_{seq}$ and the probability distribution output of sequence task as $p_{\theta_{seq}}(\cdot|\bi{x})$ in the following discussions. Finally, we use a well-known policy gradient approach named Proximal Policy Optimization(PPO) \cite{schulman2017proximal}, which has achieved great success recently.
\begin{eqnarray*}\label{3}
&L_{seq}(\theta_{seq}) = \mathbb{E}_t[\min(r_t(t)\hat{A},clip(r_t(t),1-\epsilon,1+\epsilon)\hat{A}) + \hat{A}^2]
\end{eqnarray*}
Here, $r_t(t)=\frac{\pi_{\theta_{seq}}(s_t|\bi{x})}{\pi_{\theta_{seq}^{old}}(s_t|\bi{x})}$ denotes the probability ratio between the updated policy parameters $\theta_{seq}$ and the vector of policy parameters before update $\theta_{seq}^{old}$, $\hat{A}=-SA(\bi{s},\bi{o}|\bi{x})-V(\bi{x})$ denotes an estimator of the advantage function, $SA(\bi{s},\bi{o}|\bi{x})$ denotes the surface area of the bin in the case of placement sequence and its corresponding orientation of $\bi{o}$ and $\bi{s}$. Notably, the reward function is the negative of $SA(\bi{s},\bi{o}|\bi{x})$, as the smaller $SA(\bi{s},\bi{o}|\bi{x})$ is better.

\subsection{Orientation Task}
As mentioned before, there are 6 orientations for each cuboid-shaped item. For orientation generation, the decoder computes the orientation based on a context vector $\bi{c}$, which describes the whole problem and is based on the partial solution constructed so far. That is the context of the decoder at time step $t$ comes from the encoder outputs and the outputs up to time step $t$. For each time step $t$, we calculate the context vector:
\begin{eqnarray*}\label{4}
 	c_t = [h^{e};h_{t}^{d};h^{intra}_{t}].
\end{eqnarray*}
Here, [.;.;] denotes vector concatenation,
$h^{e}$ denotes the representation of input sequence and is calculated by an attention-polling function which is defined as follows:
\begin{eqnarray*}
	&h^{e} = \sum_{j=1}^{n}{attc_{tj}^{d} * h_{j}^e}, \\
	&attc_{tj}^{d} = softmax({v_{3}}^{T}tanh(W_{6}h_{j}^{e} + W_{7}{[h_{t}^{d};h^{intra}_{t}]})), \\
	& j \in \{1,\dots,n\}.
\end{eqnarray*}
Where $W_{6}$, $W_{7}$ and $v_{3}$ are trainable parameters for attention-polling function.
And we apply the intra-attention feature $h^{intra}_{t}$ similar to the previous section to represent the context up to current decoding step $t$.

Thus, the probability distribution of orientations for the current decoding step $t$ is generated by as follows:
\begin{eqnarray*}\label{5}
	p(o_{t}) = softmax(W_{ori}c_t + b_{ori}),
\end{eqnarray*}
where $W_{ori}$ and $b_{ori}$ are trainable parameters for orientations.

We define $\bi{o}^{*} = \{o_{1}^{*},o_{2}^{*}, ..., o_{n}^{*}\}$ as the ground-truth orientation sequence for a given input $\bi{x}$ and generated packing item sequence $\bi{s}$. Thus the orientation task parameterized by $\theta_{ori}$ can be trained with the following standard differentiable loss (the cross-entropy loss):
\begin{eqnarray*}
	L_{ori}(\theta_{ori}) = -\sum_{i=1}^{n}\log{p(o_{i}^{*}|s_{1}, o_{1}^{*},..., s_{i-1},o_{i-1}^{*}, \theta_{ori}, \bi{x})}.
\end{eqnarray*}

Inspired by the Hill Climbing (HC) algorithm \cite{russell2016artificial}, which starts with an arbitrary sample solution to a 3D-FBPP and then attempts to find a better solution by making an incremental change to the solution, we will always keep the best solution for each problem instance, and train on it and use the new model to generate a better one.
In that case, the orientation in the current best solution is the ground-truth orientation $\bi{o}^{*}$.

\subsection{Training}
\label{sec:train}
In this subsection, we will illustrate the training  which is specially designed for the 3D-FBPP.
To train a multi-task model, we use a hybrid loss function to combine supervised learning and reinforcement learning process. 
Nevertheless, as the orientation task is much more complex(For example, if we have 8 items to be packed, the choices of \emph{Sequence Generation} is $8!=40320$ and the \emph{Orientation Generation} is $6^8=1679616$), the sequence task will suffer from the bad initializer of orientation if they are trained at the same time. To overcome this shortcoming, pre-training the orientation task first could be a reasonable idea.
However, orientations of the items is tightly attached to the existence of packing sequence of items in the 3D-FBPP. 
Consequently, as a trade-off, we train different types of tasks separately in each batch to keep them dynamic balance.
We called this \textbf{M}ulti-\textbf{t}ask \textbf{S}elected \textbf{L}earning (MTSL). Mathematically, there are three kinds of basic loss function in our work: $L_{seq}$, $L_{ori}$ and $L_{all}$.
\begin{eqnarray*}
	&L_{all}  = \alpha* L_{seq}(\theta_{seq}) +(1-\alpha)* L_{ori}(\theta_{ori}).
\end{eqnarray*}
Where $\alpha$ is the hyper-parameter we will fine-tune in our experiments.
During training, to utilize correlation and relieve imbalance, our model will choose one of the three kinds of losses $\{L_{seq}, L_{ori}, L_{all}\}$ according to a probability distribution which will decay after several training step.

As a conclusion of the discussion above, the training procedure of our Multi-task Selected Learning method is summarized in Algorithm \ref{alg-summary-drl}.

\begin{algorithm}                  
	\caption{Multi-task Selected Learning.}       
	\label{alg-summary-drl} 
	\begin{algorithmic}[1]
		\State Training set $\bi{X}$, training steps $T$, batch size $B$.
		\State Init. Sequence and Orientation parameters $\theta_{seq}$, $\theta_{ori}$.
		\State Init. best solution pool $D = \varnothing$
		\For {\emph{t = 1 to T}}
		\State Select a batch of sample $\bi{x}$.
		\State Sample a sequence $\bi{s}$ according to prob. $P_{\theta_{seq}}(\cdot|x)$.
		\For {\emph{m = 1 to k}}
		\State Sample orientations $\bi{o}$ according to prob. $P_{\theta_{ori}}(\cdot|x)$.
		\State Obtain empty-maximal-space $\bi{f}$ by LWSC.
		\State Calculate $SA(\bi{s},\bi{o}|\bi{x})$ by tuple $(\bi{s},\bi{o},\bi{f})$
        \EndFor
        \State keep the current best solution tuple $(\bi{s},\bi{o},\bi{f},\bi{x})$
        \State Compare with the best solution pool D and get the best $(\bi{s},\bi{o}^{*},\bi{f},\bi{x})$ for $\bi{x}$
        \State update D with the best solution $(\bi{s},\bi{o}^{*},\bi{f},\bi{x})$
		\State Calculate sequence and orientation task gradient $g_{\theta_{seq}}$, $g_{\theta_{ori}}$ based on the tuple $(\bi{s},\bi{o}^{*},\bi{f},\bi{x})$ 
		\State  $g_{\theta} = choice(g_{\theta_{seq}},g_{\theta_{ori}},g_{\theta_{all}})$
		\State Update $\theta = ADAM(\theta, g_{\theta})$.
		\EndFor
		\State return all parameters $\theta_{seq}$, $\theta_{ori}$.
	\end{algorithmic}

\end{algorithm}

\section{Experiments}
\label{sec:exp}

We conduct experiments to investigate the performance of the proposed \textbf{M}ulti-\textbf{t}ask \textbf{S}elected \textbf{L}earning (MTSL) 3D-FBPP methods. As mentioned above, our experiments are conducted on the proposed \textbf{L}arge-scale \textbf{R}eal-world \textbf{T}ransaction \textbf{O}rder \textbf{D}ataset (LRTOD).

\subsection{Dataset Description}
\label{sec:data}
We collect a dataset on 3D flexible bin packing problem based on real-world order data from Taobao's supermarket warehouse management system. This LRTOD consists of two parts: the customer order data from E-commerce platform and its corresponding items' size data (i.e., length, width and height) from Logistics platform. As in real e-commerce scenario, 83.6\% of the orders contain statistically less than 10 items and only 0.2\% of them over 40, so the problem we studied are relative smaller than the dataset of 3DBPP generated by a well-known released code \footnote{\url{http://hjemmesider.diku.dk/~pisinger/codes.html}}. In particular, we randomly sample 150,000 training data and 150,000 testing data from customer orders with 8, 10 and 12 items, which are named as BIN8, BIN10 and BIN12 respectively. We believe this real-world dataset will contribute to the future research of 3DBPP. \footnote{The data will be published soon after accepted.}

\subsection{Implementation Details}
\label{sec:implementation}
Across all experiments, we use mini-batches of fixed size 128 and LSTM cells with 256 hidden units. In addition, description of each item's size is embedded into a 256-dimension input. We train our model with Adam optimizer \cite{kingma2014adam} by initial learning rate of $10^{-3}$ and decay every 5000 steps by a factor of 0.96. 
For hyper-parameter $\alpha$ in loss function $L_{all}$, we fine-tune it and set $\alpha=0.5$.
We use the clipped surrogate objective of PPO and the corresponding hyper-parameter $\epsilon$ is 0.2. We use 1000,000 steps to train the model and it will take about a few hours on Tesla M40 GPU machine. Model implementation with TensorFlow will be made available soon.
Based on comprehensive consideration, the performance indicator is average surface area (ASA) which denotes the average cost of packing materials. The mathematical definition of ASA is $ \frac{\sum_{i=1}^{n} SA(i)}{n}$, 
where $SA(i)$ is the surface area of $i$th order and $n$ is the number of orders. We will show the compared methods and detailed experiments in the following.

\subsubsection{Single Task Pointer Network}
Pointer network used in TSP is to generate the sequence of the cities to visit, which lays the foundation for learning solutions to problems using a purely data-driven approach. We use the Pointer network to produce the placement sequence of items in a 3DFBPP and other tasks such as $\emph{Orientation Generation}$ and $\emph{Spatial Location Selection}$ are finished by LWSC mentioned above. The main difference between TSP and 3D-FBPP is the input and reward function. In our setting, the input is the width, height and length of items and the reward function is the surface area of the packed bin. Other experimental settings substantially coincide with ones solving TSP. We refer the method which never takes the packing sequence that has already been generated into consideration as RL-vanilla in our paper. As a contrast, method which introduces the intra-attention mechanism are refered as RL-intra.
It is worth mentioning that about other RL models in the following sections have introduced the intra-attention mechanism by default. 

\subsubsection{Multi-task Selected Learning}
\label{sec:mtsl-train}
For MTSL experiments, at each step, we choose a loss function of $\{L_{seq}, L_{ori}, L_{all}\}$ (in \secref{sec:train}) dynamically adapting to the training process. Actually, we sample these three losses with probabilities $(0.3, 0.5, 0.2)$. The values of probabilities are annealed to $(0.33, 0.33, 0.33)$ respectively over the first 10,000 training steps. Similar to evaluation of RL-vanilla and RL-intra
, the results of MTSL are obtained by beam search with beam size $5$ as in Table \ref{rl-result}. 

Most heuristic methods usually search the solution space in parallel and compare the candidate solutions. It is enlightening to achieve better results using a sampling framework that  constructs multiple solutions and chooses the best one. Based on this, we also introduce sampling into our proposed methods, which sample 128 solutions for each test instance and report the least surface area. 
For comparison, we compare three different methods: 1) the sequence is generated by the RL-intra model and orientation is generated according to LWSC; 2) the sequence and orientation are both generated by our MTSL model; 3) the sequence and orientation are both generated by the multi-task model without Selected Learning whose loss function is $L_{all}$. We refer to those results as RL-intra-Sample, MTSL-Sample and MT-Sample. These added results are shown in Table \ref{bin result}. Naturally, the difference with other approaches is diluted by sampling many solutions (Even a random policy may sample good solutions occasionally). 

\subsection{Results and Analysis}
\label{sec:eval}
First of all, we evaluate different models including RL-vanilla, RL-intra and MTSL on our proposed dataset LRTOD. We report the ASA results in Table \ref{rl-result}. The problem cannot be solved directly by optimization solvers, such as Gurobi \cite{optimization2014inc}, because its Hessian matrix is not positive or semi-positive definite. As the table shows, RL-vanilla achieves $4.89\%$, $4.88\%$, $5.33\%$ improvement than LWSC for BIN8, BIN10 and BIN12, whereas the improvement of RL-intra is increased to $5.19\%$, $5.26\%$, $5.41\%$ respectively. Apparently, it demonstrates the usefulness of our intra-attention training mechanism, which can help reduce the surface area of the 3D-FBPP. Moreover, the significant results of MTSL also show that the orientation distribution trained by MTSL model comfortably surpass a greedy orientation strategy produced by LWSC. Overall, MTSL obtains 6.16\%, 9.66\%, 8.25\% surface area reduction than the well-designed heuristic LWSC. 

In order to improve the adequacy of contrast test, we also conduct the approach BRKGA in \cite{gonccalves2013biased} on LRTOD to validate whether these methods designed for fixed-sized bin are appropriate for our 3D-FBPP. BRKGA is one of the state-of-the-art methods to tackle 2D and 3D fixed-sized bin packing problems which adopts a heuristic method of Genetic Algorithm (GA) \cite{whitley1994genetic}. For fair comparison, we first change the objective from minimizing the number of bins to finding a minimized surface area bin. In BRKGA, the spatial location strategy Distance to the Front-Top-Right Corner (DFTRC) is rather sensitive to the given bin size and the main issue of our problem is the flexible-sized bin. To achieve a good solution, we test BRKGA with different sized bins by grid search. Besides, we also adopt a spatial location strategy which rarely depends on the bin size and usually cooperates with hybrid GA named deepest bottom left with fill (DBLF) \cite{Karabulut2004A, Wang2010A} to replace DFTRC. Moreover, to verify the effectiveness of LWSC, we also use LWSC as the spatial location strategy to compare with the heuristic methods mentioned above. On one hand, according to Table \ref{rl-result}, we find that GA+LWSC works very well and use GA to evolve the packing sequence and orientation is rather effective. On the other hand, the better results of MTSL illustrate that our model can generate better bin packing sequence and orientation than GA in saving packing cost. To sum up, the statistical analysis confirms that MTSL is significantly better than all the other approaches outlined above.
\begin{table}[!htb]
	\caption{Comparison with RL-vanilla and RL-intra on the LRTOD. The difference among GA+LWSC, GA+DBLF, BRKGA is the spatial location strategy and BRKGA is the only method which use grid search to obtain a better solution. Note that the unit of the surface area in all experiments is $cm^2$. }
	\centering
	\begin{tabular}{clll}
		\toprule	
		model &   BIN8 & BIN10 & BIN12 \\ \midrule
		Random &  44.70 & 48.38 & 50.78 \\ 
		Gurobi &   --   & --    & --  \\
		LWSC & 43.97 & 47.33 & 49.34 \\
		BRKGA (GA+DFTRC)   &43.44 & 47.84 & 50.01 \\
		GA+LWSC &  42.44   &   44.49    &  48.77     \\
		GA+DBLF & 42.22    &  46.87  &  50.70      \\
		RL-vanilla & 41.82 & 45.02 & 46.70 \\ 
		RL-intra & 41.69 &  44.84 & 46.67 \\
		MTSL & \textbf{41.26} & \textbf{42.76} & \textbf{45.27} \\\bottomrule
	\end{tabular}
    \label{rl-result}
\end{table}

\begin{figure*}[!htb]
	\centering
	\epsfig{file=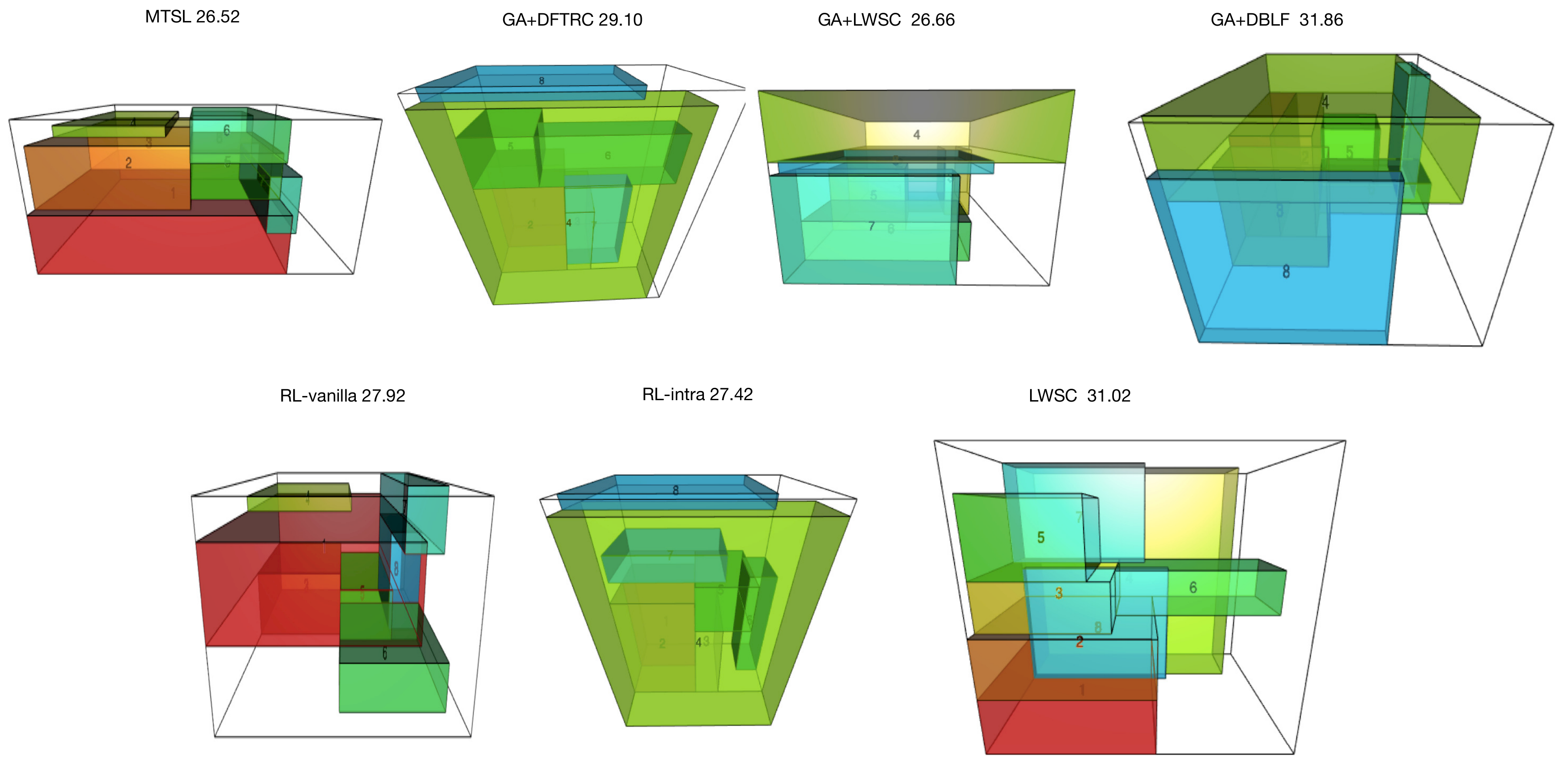, width=1.6\columnwidth}
	\caption{Results of MTSL-vanilla, GA+DFTRC, GA+LWSC, GA+DBLF,  RL-vanilla, RL-intra and LWSC. The surface area of these method are 26.52, 29.1, 26.66, 31.86, 27.92, 27.42 and 31.02 respectively}
	\label{fig:result-example}
\end{figure*}

We also report the ASA results of multiple methods which introduce sampling mechanism in Table \ref{bin result}. By calculation, the proposed MTSL-sample achieves $6.21\%$, $10.06\%$, $8.55\%$ improvement than the greedy LWSC for BIN8, BIN10, and BIN12. In our experiments, MTSL-sample is superior than other methods in most cases but is slightly less competitive than RL-intra-sample in BIN8. The comparisons between MT-sample and MTSL-sample also indicates the effectiveness of Selected Learning. Without Selected Learning, the multi-task method performs much worse than the single task method RL-intra-sample. 
\begin{table}[!htb]
	\caption{ Multiple solutions experiment. Best of 128 sampled solutions in RL-intra, multi-task without Selected Learning and MTSL model. }
	\centering
	\begin{tabular}{clll}
		\toprule	
		model &   BIN8 & BIN10 & BIN12 \\
		\midrule
		LWSC & 43.97 & 47.33 & 49.34 \\ 
		RL-intra-sample &\textbf{41.12} & 44.03 & 45.58 \\ 
		MT-sample & 42.31 & 45.01 & 45.62 \\
		MTSL-sample & 41.24 & \textbf{42.31} & \textbf{45.12} \\ 
		\bottomrule
	\end{tabular}
	\label{bin result}
\end{table}

Finally, we randomly choose an example order instance from BIN8 and display packing results by different methods in Figure \ref{fig:result-example} for case study. As shown, the surface area computed by each method is listed on the top of its corresponding image. Obviously, the packing results demonstrate that MTSL can produce a more reasonable packing policy than other methods. 

\subsection{Online Experiment}
\label{sec:real}
Having obtained encouraging results on the large-scale real-world transaction order dataset, we finally perform our proposed method on our online production system of Taobao. We first detail the design of our experiments, and then show its corresponding results.

\subsubsection{A/B Test Design.} A/B testing is a widely-used method for comparing two or more varied algorithms in real-world systems, including item recommendation and real-time bidding. We first deploy our method on online supermarket warehouse system of Taobao across 4 different cities of China and then we do a traffic assigning strategy that randomly splits users into 50/50 groups for LSWC and MTSL.

\subsubsection{Experiment Setup and Results.} Our proposed method and its comparison method has undertaken online A/B testings for one month . MTSL is initially trained based on historical data from previous real production data. We use the same hyper-parameters as mentioned in Section \ref{sec:implementation}. Considering sensitivity of business data, experimental results ignore the real package cost. Table \ref{online-test} show that the performance improvement brought by our method is consistent in all cities, with gains in
global cost reduction rate ranging from $5.5\%$ to $6.6\%$. Given these promising results, the proposed algorithm has been successfully deployed in Taobao's supermarket warehouse system system for more than 20 major cities, saving a huge logistics cost in a daily basis.

\begin{table}[!htb]
	\caption{Comparison of cost reduction results on online A/B testings in four cities. Rate stands for the average cost reduction rate for orders.}
	\centering
	\begin{tabular}{cl}
		\toprule
		City &   Rate \\ \midrule
		City A &  $-5.7\%$   \\ 
		City B &  $-6.6\%$  \\ 
		City C &  $-5.1\%$ \\ 
		City D &  $-4.5\%$ \\ 
		\bottomrule
	\end{tabular}
	\label{online-test}
\end{table}

\section{Conclusion}
\label{sec:conclusion}

In this paper, we first introduce, define and solve the 3D Flexible Bin Packing Problem. We model this problem as a sequential decision-making problem and propose a multi-task framework based on selected learning to generate packing sequence and orientations simultaneously, which can utilize correlation and mitigate imbalance between the two tasks. We also adopt an intra-attention mechanism to address the repeated item problem and use the idea of hill-climbing algorithm to learning a more competitive orientation strategy. Through comprehensive experiments, we achieves $6.16\%$, $9.66\%$, $8.25\%$ improvement than the greedy algorithm designed for the 3D-FBPP in BIN8, BIN10 and BIN12 and an average $5.47\%$ cost reduction on online AB tests across different cities' supermarket warehouse system of Taobao. Numerical results also demonstrate that our selected learning method significantly outperforms the single task Pointer Network and the multi-task network without selected learning. A large-scale real-world transaction order dataset is collected and will be released after company's internal audit. In future research, we will focus on investigation of more effective network architecture and learning algorithm. Meanwhile, it is beneficial to apply our proposed method to more interesting combinatorial optimization problems in the domain of logistics to help reduce costs of the industry.


\bibliographystyle{ACM-Reference-Format}  
\bibliography{ref}  

\end{document}